# MSC-LIO: An MSCKF-Based LiDAR-Inertial Odometry with Same-Plane Cluster Tracking

Tisheng Zhang, Man Yuan, Linfu Wei, Hailiang Tang*, and Xiaoji Niu

***Abstract*—The multi-state constraint Kalman filter (MSCKF) has been proven to be more efficient than graph optimization for visual-based odometry while with similar accuracy. However, it has not been adequately considered and studied for LiDAR-based odometry. In this paper, we propose a novel tightly-coupled LiDAR-inertial odometry based on the MSCKF framework, named MSC-LIO. An efficient LiDAR same-plane cluster (LSPC) tracking method, without explicit feature extraction, is present for frame-to-frame data associations. The tracked LSPC is used to build an LSPC measurement model that constructs multi-state constraints. Besides, we propose an effective point-velocity-based LiDAR-IMU time-delay (LITD) estimation method, which is derived from the proposed LSPC tracking method. To validate the effectiveness and robustness of the proposed method, we conducted extensive experiments on both public datasets and real-world environments. The results demonstrate that the proposed MSC-LIO yields higher accuracy and efficiency compared to the state-of-the-art methods. Ablation experiments indicate that the data-association efficiency is improved by nearly 3 times with the LSPC tracking, and the proposed LITD estimation method can effectively and accurately estimate the LITD. Besides, MSC-LIO was implemented on an edge device and demonstrated excellent real-time performance.**

***Index Terms*—State estimation, light detection and ranging (LiDAR)-inertial odometry, multi-state constraint Kalman filter (MSCKF), multi-sensor fusion navigation.**

## I. Introduction

LIGHT detection and ranging (LiDAR) has played an increasingly important role in autonomous vehicles and robots, mainly due to its spatial perception capabilities and the rapid development of low-cost solid-state LiDARs. Meanwhile, the inertial measurement unit (IMU) can be employed for navigation independently, and the inertial navigation system (INS) can output high-frequency poses [1]. Hence, the INS can be adopted to correct the motion distortion of point clouds. For this reason, the LiDAR and the IMU have been integrated to construct the LiDAR-inertial odometry (LIO) for more accurate pose estimation. LIOs can be categorized as optimization-based and filter-based depending on the type of the state estimator. Generally, graph optimization provides higher accuracy than filters but at the cost of lower efficiency [2]. The multi-state constraint Kalman filter (MSCKF) [3] has been proven to achieve similar accuracy to graph optimization while more efficient in visual-inertial odometry (VIO) [4], [5]. Recently, some studies [6], [7], [8] have explored the application of MSCKF in LIOs and proved that the MSCKF achieves higher accuracy than other filters for LiDAR-inertial navigation. Nevertheless, the data-association method and the LiDAR measurement model in MSCKF-based LIOs should still be further studied for efficient and accurate state estimation.

This research is partly funded by the National Natural Science Foundation of China (No.42374034), the Key Research and Development Program of Hubei Province (2024BAB024), the Major Program (JD) of Hubei Province (2023BAA02602), and the High quality development project of MIIT (2024-182). (*Corresponding author: Hailiang Tang.*)

Tisheng Zhang and Xiaoji Niu are with the GNSS Research Center, Wuhan University, Wuhan 430079, China, and also with the Hubei Luojia Laboratory, Wuhan 430079, China (e-mail: zts@whu.edu.cn; xjniu@whu.edu.cn).

Man Yuan and Hailiang Tang are with the GNSS Research Center, Wuhan University, Wuhan 430079, China and also with the School of Geodesy and Geomatics, Wuhan University, Wuhan 430079, China (e-mail: yuanman@whu.edu.cn; thl@whu.edu.cn).

Linfu Wei are with the GNSS Research Center, Wuhan University, Wuhan 430079, China (e-mail: weilf@whu.edu.cn).

### A. Related Works

#### 1) Optimization-Based Methods

Some studies employ frame-to-map (F2M) association methods for LiDAR. LIO-mapping [9] and GIVL-SLAM [10] built a local feature map within a sliding window and matched LiDAR features (edges and planes) to the map. LIO-SAM [11] built a voxel map for edge and planar features and applied a sliding window-based F2M scan-matching method to derive LiDAR measurements. However, F2M association may produce incorrect absolute measurements, resulting in inconsistent state estimation [12]. In contrast, frame-to-frame (F2F) association constructs relative measurements and ensures consistent state estimation. LIOs based on F2F association can seamlessly integrate with absolute positioning sensors, such as the global navigation satellite system (GNSS) [13] and ultrawideband (UWB) [14]. Thus, some studies on LIO have shifted their focus to frame-to-frame (F2F) association methods. LIPS [15] proposed an anchor plane factor that associates multiple LiDAR frames. VILENS [16] tracked plane and line features across multiple frames and constructed corresponding landmark factors. BA-LINS [17] developed an F2F bundle adjustment (BA) measurement model for LiDAR. The BA measurements were derived from same-plane points in the keyframes within a sliding window. The same-plane points were obtained through direct point-cloud processing [12], [18], which is more efficient than the feature extraction methods used in [9], [10], [11], [15], and [16].

The above methods rely on optimization frameworks, which require multiple iterations and suffer from low efficiency. In contrast, filters offer higher efficiency but at the cost of reduced accuracy. However, unlike VIOs, LIOs can avoid estimating extra landmark states that introduce significant linearization errors. Thus, filters should theoretically achieve comparable accuracy to optimization-based methods for LIO.

2) *Filter-Based Methods*

Filters typically exhibit lower accuracy compared to graph optimization, primarily due to linearization errors. The iterated error-state Kalman filter (IESKF) mitigates linearization errors by iterative updates, ensuring both accuracy and efficiency [19], which has made it widely adopted in numerous studies. FAST-LIO [20] extracts edge and planar features and employs the IESKF to tightly integrate IMU and LiDAR measurements, achieving greater efficiency compared to optimization-based LIO-SAM [11]. To address the growing computation issue of FAST-LIO, FAST-LIO2 [18] directly registers raw LiDAR points to a global map represented by an incremental k-d tree and employs the IESKF for accurate state estimation. The probabilistic adaptive voxel map proposed in [21] consists of voxels that each represent a plane feature and is integrated into the IESKF framework. VoxelMap++ [22] expands on the work in [21] by consolidating coplanar features into larger planes, thereby reducing uncertainty in the overall map and improving the state estimation accuracy of the IESKF. The IESKF-based LIOs discussed above rely on F2M association (i.e., matching the current frame to the global map) instead of F2F association for higher accuracy. The reason is that the IESKF estimates only the current pose, and relying solely on F2F association between two consecutive frames makes it challenging to achieve high accuracy. However, F2M association may produce incorrect absolute measurements, hindering the seamless integration of absolute positioning sensors [12], [23].

MSCKF can also achieve high accuracy by simultaneously estimating multiple poses within a sliding window, without the need for iterations. *A key advantage of MSCKF over IESKF is its ability to construct multi-state relative constraints, enabling it to maintain high accuracy with F2F association.* LIC-Fusion [7] efficiently fuses IMU measurements and LiDAR features (planes and edges) within the MSCKF framework. Extracted features from the current LiDAR frame are tracked back to the previous frame to construct the F2F measurement model. However, LIC-Fusion constructs the F2F associations only between two LiDAR frames, without fully leveraging the multi-state characteristics of MSCKF. In contrast, LIC-Fusion 2.0 [8] adopts a normal-based method to associate plane features and stores them as SLAM plane landmarks. The plane landmarks can be associated with multiple frames, implementing multi-state F2F constraints and thus achieving higher accuracy. Additionally, LIC-Fusion 2.0 proposes a plane-feature tracking method to enhance efficiency.

The multi-state relative constraints in BA-LINS [17] and LIC-Fusion 2.0 [8] resemble the bundle adjustment (BA) in visual multiple-view geometry [24], which has been proven to be more accurate in LiDAR mapping by BALM [25]. However, BALM is designed for LiDAR mapping rather than odometry. BA-LINS is an optimization-based method suffering from poor efficiency. Although LIC-Fusion 2.0 utilizes the MSCKF framework, its reliance on explicit plane-feature extraction increases computational costs notably and leads to poor accuracy in unstructured environments. As a result of explicit plane-feature extraction, LIC-Fusion 2.0 augments plane landmarks with a long track length into the state vector, resulting in lower state-estimation efficiency. Moreover, the efficient plane-feature tracking proposed in LIC-Fusion 2.0 is designed specifically for spinning LiDARs and not applicable to the increasingly popular solid-state LiDARs. Hence, the high-efficiency potential of the MSCKF remains underutilized in existing LIOs, and the LiDAR measurement model should be further studied for accurate MSCKF state estimation.

### B. Main Contributions

Motivated by the above issues, this study proposes a tightly-coupled MSCKF-based LIO with LiDAR same-plane cluster (LSPC) tracking, named MSC-LIO. Here, the proposed LSPC tracking builds on the point-to-plane Iterative Closest Point (ICP) [26]. Each LiDAR point is evaluated with the point-to-plane ICP to determine whether it qualifies as a plane point. *When a point is identified as a plane point, it is paired with its nearest neighbor, and they are referred to as a cluster of points on the same plane. The process of associating LiDAR same-plane clusters between two consecutive LiDAR frames is defined as LSPC tracking, which is proposed to improve data-association efficiency.* To fully exploit the characteristics of MSCKF, same-plane clusters derived from LSPC tracking, which establish relative constraints across multiple LiDAR frames, are employed to construct the LSPC measurement model. The main contributions of this study are as follows:

1) We present an MSCKF-based LIO that tightly integrates IMU and LSPC measurements, with online calibration of LiDAR-IMU spatiotemporal parameters.

2) We propose an efficient LSPC tracking method that avoids explicit plane-feature extraction and is compatible with both spinning and solid-state LiDARs. By tracking LSPCs frame by frame within a sliding window, this method significantly enhances data-association efficiency.

3) By leveraging LSPC tracking, the velocity of plane points can be approximately calculated. Using the point velocity, we propose a novel and straightforward approach to estimate the LiDAR-IMU time delay (LITD).

4) Comprehensive experiments are conducted on both public datasets and real-world environments to evaluate the performance of the proposed MSC-LIO. Results demonstrate that our method outperforms the state-of-the-art (SOTA) methods in both accuracy and efficiency.

## II. System Overview

The system overview of the proposed MSC-LIO is shown in Fig. 1. We adopt an INS-centric processing pipeline. The initial position and heading angle of the IMU are initialized to zero, while the roll and pitch angles are determined by the accelerometer measurements and gravity acceleration. If a zero-velocity state is detected during initialization, we can also obtain an initial gyroscope bias using the mean gyroscope measurements. Once the system is initialized and the IMU input is received, the INS mechanization is performed, and the state vector and its covariance are forward propagated. In the meantime, the INS pose is stored for further point-cloud preprocessing. When a LiDAR frame is received, the point-cloud distortion is corrected with the high-frequency INS pose. Besides, the INS prior pose is also employed for keyframe selection. The keyframe strategy is employed to balance accuracy and efficiency [11]. If the relative motion or time interval between the current frame and the previous keyframe exceeds the predefined threshold, the current frame is considered a keyframe [12]. The keyframe-selection thresholds are detailed in Section V.A.

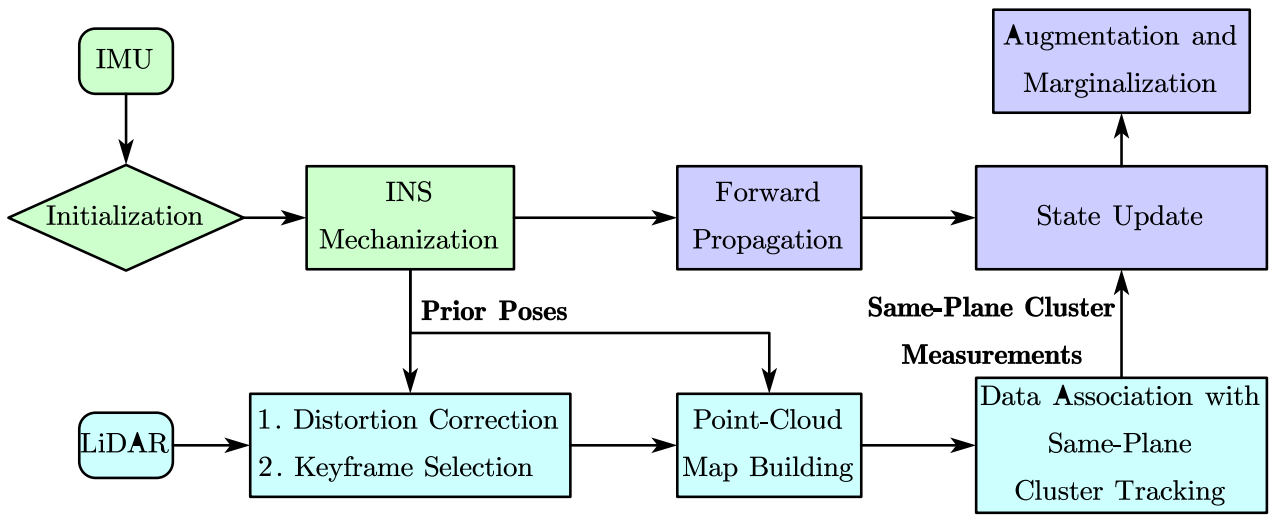

Fig. 1. The system overview of the proposed MSC-LIO.

When a keyframe is selected, all non-keyframes between the previous keyframe and the current keyframe are projected and merged into the current keyframe to construct the keyframe point-cloud map. Subsequently, the LSPC tracking is performed between the historical keyframe point-cloud maps and the current. More specifically, the LSPC candidates in the historical keyframe point-cloud maps are utilized to track the LSPCs in the current keyframe point-cloud map. The LSPC tracking is conducted by searching for nearest neighboring points instead of explicit plane feature extraction. The tracked LSPCs are employed to construct the F2F LSPC measurement model. Then, the LSPC measurements are utilized to update the IMU state, the keyframe states, and the LiDAR-IMU spatiotemporal parameters within the MSCKF framework. Finally, the current keyframe pose state is augmented into the MSCKF state vector. The marginalization is conducted when the sliding window exceeds its maximum length.

## III. MSCKF-Based LiDAR-Inertial Estimator

### A. State Vector

The error state vector of the MSCKF includes the IMU state $\delta\boldsymbol{x}_I$, the LiDAR-IMU extrinsic parameter state $\delta\boldsymbol{x}_l^b$, the LiDAR-IMU time delay state $\delta t_d$, and the keyframe states $\delta\boldsymbol{x}_k (k=0,1,\cdots,N-1)$, where $N$ is the length of the sliding window. The error state $\delta\boldsymbol{x}$ can be written as

$$\delta\boldsymbol{x} = [\delta\boldsymbol{x}_I, \delta\boldsymbol{x}_l^b, \delta t_d, \delta\boldsymbol{x}_0, \delta\boldsymbol{x}_1, \cdots, \delta\boldsymbol{x}_{N-1}]^T, \tag{1}$$

where $\delta\boldsymbol{x}_I$, $\delta\boldsymbol{x}_l^b$, and $\delta\boldsymbol{x}_i$ are expressed as

$$\delta\boldsymbol{x}_I = [\delta\boldsymbol{\theta}_{b_N}^w, \delta\boldsymbol{p}_{b_N}^w, \delta\boldsymbol{v}^w, \delta\boldsymbol{b}_g, \delta\boldsymbol{b}_a], \tag{2}$$

$$\delta\boldsymbol{x}_l^b = [\delta\boldsymbol{\theta}_l^b, \delta\boldsymbol{p}_l^b], \tag{3}$$

$$\delta\boldsymbol{x}_k = [\delta\boldsymbol{\theta}_{b_k}^w, \delta\boldsymbol{p}_{b_k}^w]. \tag{4}$$

Here, $W$, $b$, $l$ represent the world frame, the IMU frame, and the LiDAR frame, respectively; $\delta\boldsymbol{\theta}_{b_N}^W$, $\delta\boldsymbol{p}_{b_N}^W$ and $\delta\boldsymbol{v}^W$ denote the errors of current IMU attitude, position, and velocity, respectively; $\delta\boldsymbol{b}_g$ and $\delta\boldsymbol{b}_a$ denote the bias errors of the gyroscope and accelerometer, respectively; $\delta\boldsymbol{\theta}_l^b$ and $\delta\boldsymbol{p}_l^b$ denote the errors of LiDAR-IMU extrinsic parameters; $\delta\boldsymbol{\theta}_{b_k}^w$ and $\delta\boldsymbol{p}_{b_k}^w$ denote the IMU pose errors at the time of keyframe $k$, which is denoted as $KF_k$. The relationship between the true state $\boldsymbol{x}$, estimated state $\hat{\boldsymbol{x}}$, and error state $\delta\boldsymbol{x}$ is

$$\boldsymbol{x} = \hat{\boldsymbol{x}} \boxplus \delta\boldsymbol{x}. \tag{5}$$

For the attitude error $\delta\boldsymbol{\theta}$, the operator $\boxplus$ is given by

$$\mathbf{R} = \hat{\mathbf{R}}\mathrm{Exp}(\delta\boldsymbol{\theta}) \approx \hat{\mathbf{R}}(\mathbf{I} + (\delta\boldsymbol{\theta})\times), \tag{6}$$

where $\mathbf{R}$ and $\hat{\mathbf{R}}$ denote the true and estimated rotation matrix, respectively; $\mathrm{Exp}$ is the exponential map [27]; $(\cdot)\times$ denotes the skew-symmetric matrix of the vector belonging to $\mathbb{R}^3$ [28]. For other states, the operator $\boxplus$ is equivalent to Euclidean addition, i.e., $\boldsymbol{a} = \hat{\boldsymbol{a}} + \delta\boldsymbol{a}$. The LITD $t_d$ is employed to adjust the LiDAR time $t_{LiDAR}$ for synchronization with IMU time $t_{IMU}$:

$$t_{IMU} = t_{LiDAR} + t_d. \tag{7}$$

When receiving the IMU measurement, INS mechanization is conducted to update the IMU pose and velocity, and the standard error-state Kalman filter (ESKF) prediction formula [29] is employed for the forward propagation of the MSCKF error state and its covariance.

### B. LiDAR Same-plane cluster Measurement Model

The tracked LSPCs are used to construct the LSPC measurement model, which is formulated by minimizing the thickness of the plane constructed by the points within each LSPC. Each LSPC contains a set of same-plane points, and each point comes from a different keyframe. The LSPC with index $j$ is denoted as $S_j = \{\boldsymbol{p}_k^l \in \mathbb{R}^3 \mid k \in \mathbb{N}_j\}$, where $\boldsymbol{p}_k^l$ represents the point belonging to $KF_k$ and $\mathbb{N}_j$ is the index set of the associated keyframes. All the same-plane points are expressed in the LiDAR frame.

Considering an LSPC $S_j$, each point is first projected onto the world frame with the LiDAR-IMU extrinsic parameters and the IMU pose as

$$\boldsymbol{p}_k^w = \hat{\mathbf{R}}_{b_k}^w(\hat{\mathbf{R}}_l^b\boldsymbol{p}_k^l + \hat{\boldsymbol{p}}_l^b) + \hat{\boldsymbol{p}}_{b_k}^w. \tag{8}$$

Then, the plane fitting is conducted with the projected points by solving an overdetermined linear equation [30]. The normalized normal vector of the plane is denoted as $\boldsymbol{n}$, and $d$ is the distance that satisfies the following equation:

$$\boldsymbol{n}^T\boldsymbol{p}^w + d = 0, \tag{9}$$

where $\boldsymbol{p}^w$ is a point on the plane in the $W$ frame. The LSPC measurement is the mean of the sum of squares of the point-to-plane distances:

$$\boldsymbol{z}^j = \frac{1}{n_j}\sum_k (\boldsymbol{n}^T \boldsymbol{p}_k^w + d)^2 \,, \tag{10}$$

where $n_j$ is the number of same-plane points in $S_j$.

### C. LiDAR Measurement Update

In the proposed method, only the LSPCs containing the point from the current keyframe are used for the update. The points in $S_j$ belong to the same plane, and thus the point-to-plane distances are zeros in the absence of errors. Hence, the residual $\boldsymbol{r}^j$ can be written as

$$\boldsymbol{r}^j = 0 - \boldsymbol{z}^j \approx \mathbf{H}_x^j \delta \boldsymbol{x} + \boldsymbol{n}_r^j \,, \tag{11}$$

where $\mathbf{H}_x^j$ is the Jacobian w.r.t to $\delta\boldsymbol{x}$, and $\boldsymbol{n}_r^j \in \boldsymbol{N}(0, \boldsymbol{\Sigma}_\Gamma^j)$ is the noise. The adaptive covariance $\boldsymbol{\Sigma}_\Gamma^j$ will be detailed in Section IV.A.

The residual $\boldsymbol{r}^j$ is the function of the poses and the LiDAR-IMU extrinsic parameters. Therefore, we can derive the corresponding analytical Jacobians using the error-perturbation method [31]. The Jacobians w.r.t the keyframe states not associated with the measurement $\boldsymbol{z}^j$ are zeros, and the Jacobians w.r.t the pose errors $\{\delta\boldsymbol{\theta}_{b_k}^w, \delta\boldsymbol{p}_{b_k}^w\}$ can be formulated as

$$\frac{\partial \boldsymbol{r}^j}{\partial \delta \boldsymbol{\theta}_{b_k}^w} = -\mathbf{J}_k^j \hat{\mathbf{R}}_{b_k}^w (\hat{\mathbf{R}}_l^b \boldsymbol{p}_k^l + \hat{\boldsymbol{p}}_l^b) \times \tag{12}$$

$$\frac{\partial \boldsymbol{r}^j}{\partial \delta \boldsymbol{p}_{b_k}^w} = \mathbf{J}_k^j, \tag{13}$$

where

$$\mathbf{J}_k^j = \frac{2}{n_j}(\boldsymbol{n}^T \boldsymbol{p}_k^w + d)\boldsymbol{n}^T \,. \tag{14}$$

Similarly, the Jacobians w.r.t the LiDAR-IMU extrinsic errors $\{\delta\boldsymbol{\theta}_l^b, \delta\boldsymbol{p}_l^b\}$ can be formulated as

$$\frac{\partial \boldsymbol{r}^j}{\partial \delta \boldsymbol{\theta}_l^b} = -\sum_k \mathbf{J}_k^j \hat{\mathbf{R}}_{b_k}^w \hat{\mathbf{R}}_l^b (\boldsymbol{p}_k) \times , \tag{15}$$

$$\frac{\partial \boldsymbol{r}^j}{\partial \delta \boldsymbol{p}_l^b} = \sum_k \mathbf{J}_k^j \hat{\mathbf{R}}_{b_k}^w \,. \tag{16}$$

The time delay $t_d$ is not considered in this section and will be discussed in Section IV.B. The standard chi-square test is utilized to remove outliers [12], and advanced methods for detecting outliers introduced by dynamic objects can also be considered [32]. With the analytical Jacobians, the MSCKF error state and covariance can be updated with the standard ESKF update formula [29].

### D. State Augmentation and Marginalization

The proposed MSC-LIO does not explicitly extract plane features, eliminating the need for null-space projection. Thus, the adopted MSCKF method only requires additional state augmentation and marginalization compared with the standard Kalman filter with fixed states. When receiving a new LiDAR keyframe, the IMU state is corrected, and then the IMU pose state is augmented into the state vector. Meanwhile, the covariance $\mathbf{P}_{n\times n}$ is augmented. The Jacobian of the augmented pose state w.r.t the state vector is $\mathbf{J}_{6\times n}$. The augmented covariance can be written as

$$\mathbf{P}_{(n+6)\times(n+6)} = \begin{bmatrix} \mathbf{I}_{n\times n} \\ \mathbf{J}_{6\times n} \end{bmatrix} \mathbf{P}_{n\times n} \begin{bmatrix} \mathbf{I}_{n\times n} \\ \mathbf{J}_{6\times n} \end{bmatrix}^T = \begin{bmatrix} \mathbf{P} & \mathbf{P}\mathbf{J}^T \\ \mathbf{J}\mathbf{P}^T & \mathbf{J}\mathbf{P}\mathbf{J}^T \end{bmatrix}. \tag{17}$$

The state and covariance of the oldest keyframe will be directly deleted when the sliding window exceeds its maximum length, that is, the marginalization [2].

## IV. LiDAR Same-Plane Cluster Tracking

The proposed LiDAR same-plane cluster tracking is based on the point-to-plane ICP [26], a direct point-cloud preprocessing method. Thus, the LSPC tracking avoids explicit feature extraction and applies to unstructured environments. The same-plane clusters are tracked between two consecutive frames instead of being searched across multiple frames [17], thereby improving data-association efficiency. Besides, based on the LSPC tracking, we propose a novel and straightforward approach to estimate the LiDAR-IMU time delay.

### A. F2F Same-plane cluster tracking Method

The proposed LSPC tracking is performed within the keyframe point-cloud maps, which are downsampled by a voxel-grid filter (with a default voxel size of 0.5 m). Since the plane features are not explicitly extracted, all points in the point-cloud maps are candidates for same-plane points. During the initialization stage, each point in the first keyframe point-cloud map is grouped into an LSPC, and LSPC tracking is subsequently performed on each LSPC. The specific procedure for the LSPC tracking is illustrated in Fig. 2. For each LSPC $S_j$, the newest point (the red points in Fig. 2) is projected onto the current keyframe point-cloud map (the yellow points in Fig. 2), and then five neighboring points (the red and blue points in Fig. 2) $\{\boldsymbol{p}_m \mid m = 1, 2, \cdots, 5\}$ are searched. The five points are used to fit a plane $(\boldsymbol{n}_l, d_l)$. If each point-to-plane distance is less than the set threshold (0.1 m), the tracking is considered valid. In this case, the nearest neighboring point (the red points in Fig. 2) is added to $S_j$ for tracking in subsequent keyframes, and the plane thickness [17] is calculated as

$$\Gamma = \frac{1}{5}\sum_{m=1}^{5} (\boldsymbol{n}_l^T \boldsymbol{p}_m + d_l)^2 \,. \tag{18}$$

For $S_j$, the covariance of the plane thickness $\boldsymbol{\Sigma}_\Gamma^j$ and the adaptive standard deviation (STD) of the point-to-plane distance $\sigma^j$ [17] can be obtained by

$$\boldsymbol{\Sigma}_\Gamma^j = \frac{1}{n_j}\sum_k (\Gamma_k^{\ 2})\mathbf{I}, \quad \sigma^j = \sqrt[4]{0.5\boldsymbol{\Sigma}_\Gamma^j} \,. \tag{19}$$

To ensure the validity of the LSPC measurements, only LSPCs containing at least 5 points and associated with the current keyframe are considered as candidates for valid measurements. The points in $S_j$ are projected onto the $W$ frame to fit a plane, and the point-to-plane distances are calculated. If all the

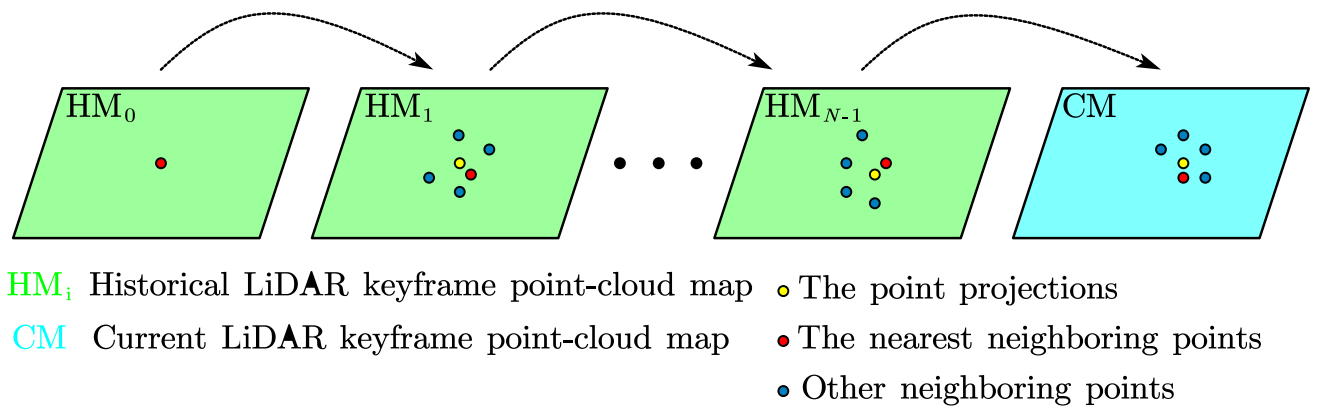

Fig. 2. An illustration of the F2F same-plane cluster tracking. Each yellow point is the projection of the red point from the previous adjacent keyframe point-cloud map.

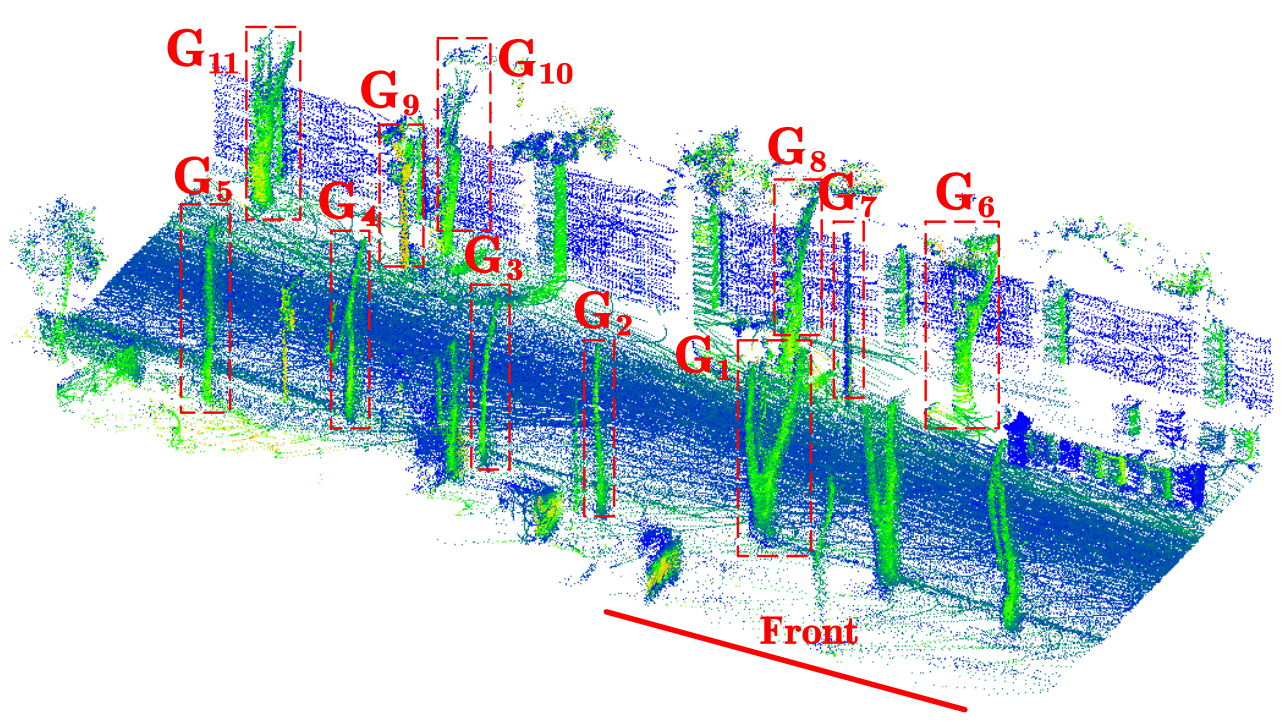

(a) The point-cloud map.

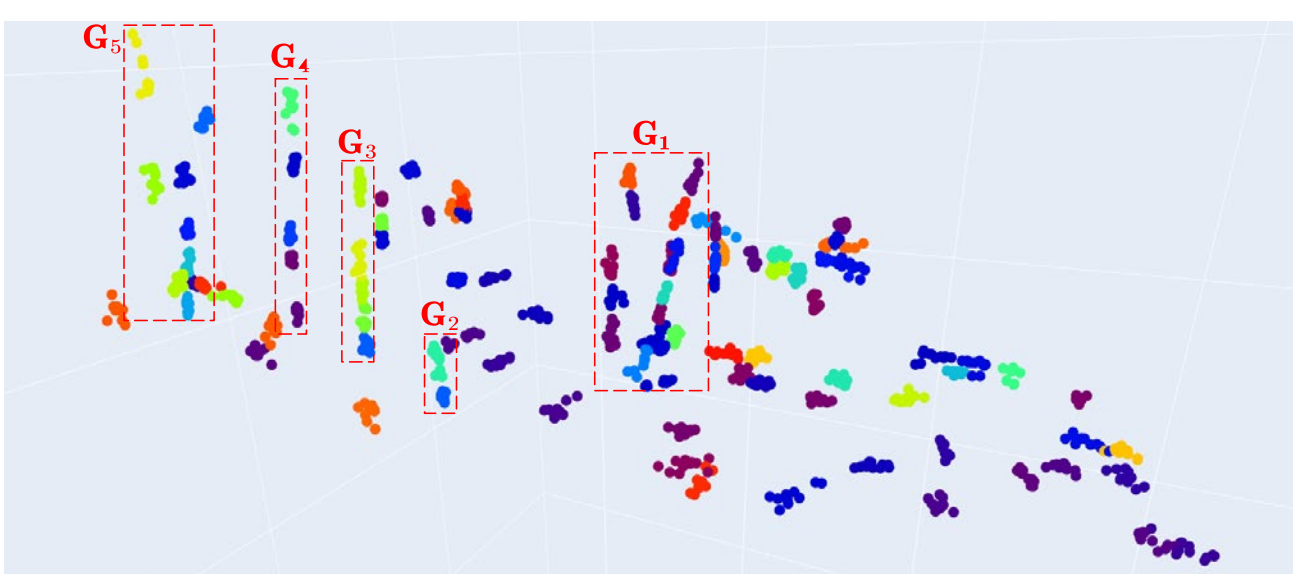

(b) Same plane clusters (left).

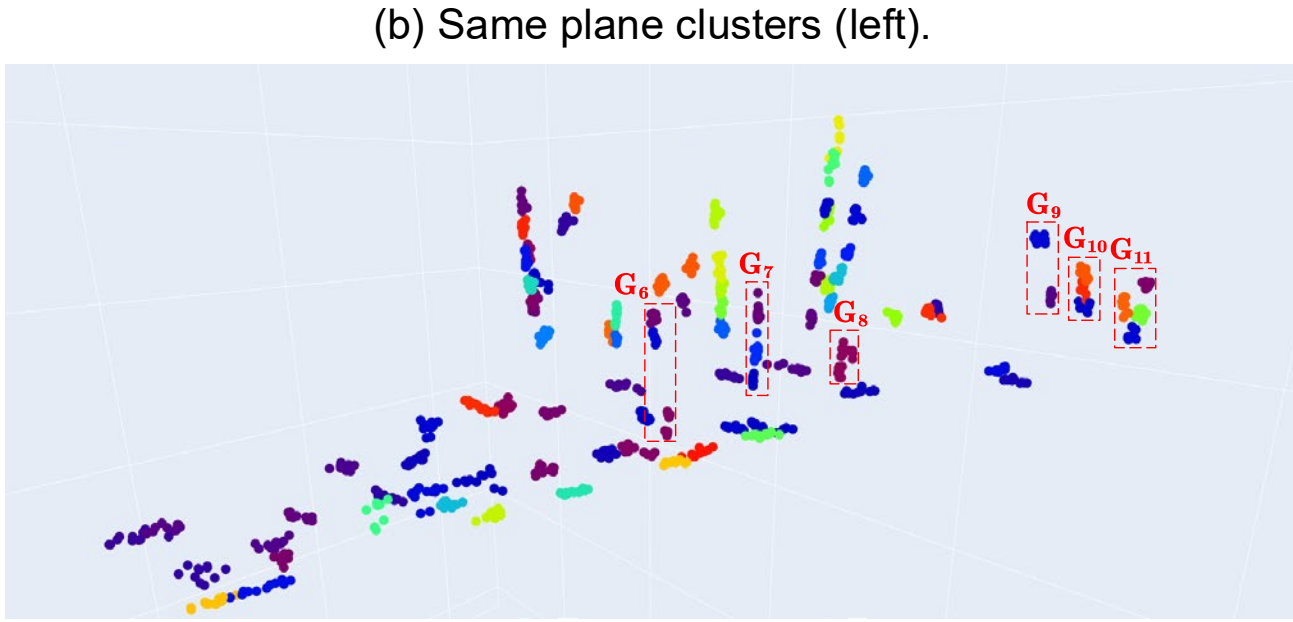

(c) Same plane clusters (right).

Fig. 3. The tracked same plane clusters. Different clusters are represented by different colors. Those clusters associated with tree trunks and other pole-like objects are marked with $G_i$, to demonstrate the effectiveness of the tracking. Each LSPC forms a small plane and the LSPC measurement model can be applied to the same-plane points within the LSPC to minimize the thickness of the plane constructed by these same-plane points. For clarity, the clusters associated with the right wall are not shown.

point-to-plane distances are less than $3\sigma^j$, $S_j$ is considered a valid measurement.

Fig. 3 shows the results of the proposed tracking method, and each LSPC is represented by a distinct color. There are many tree trunks and other pole-like objects on both sides of the road, marked with $G_i$ in the point-cloud map and tracking schematics. Most of the same-plane clusters associated with tree trunks and pole-like objects in Fig. 3 are successfully tracked, indicating the effectiveness of the LSPC tracking. Notably, only the LSPCs associated with all keyframes in the sliding window are shown in Fig. 3. For clarity, the LSPCs associated with the right wall are not shown in Fig. 3 (c).

To ensure the stability of MSC-LIO, we supplement the LSPCs with new points. The current keyframe point-cloud map is voxelized with a much larger voxel size than 0.5 m (the default is 2.5 m), and only the point closest to each voxel center is retained to ensure uniform spatial distribution of the supplementary points. The retained points are employed to construct new LSPCs to ensure the robustness of the F2F association. Besides, the clusters not associated with the newest five keyframes will be removed. An overview of the tracking method is provided in Fig. 4.

### B. LiDAR-IMU Time-Delay Estimation Based on Same-Plane Point Velocity

An unknown time delay typically exists between the timestamps of the IMU and the LiDAR. If not estimated and compensated for, the LiDAR-IMU time delay (LITD) will introduce unmodeled errors into the state estimation, thereby reducing its accuracy. Traditional LITD estimation algorithms require the angular rate and velocity to compensate for each keyframe pose [33]. Hence, we propose a simpler and more direct LITD estimation algorithm based on the LSPC tracking. The proposed algorithm utilizes the point velocity exclusively to compensate for the positions of LiDAR same-plane points.

The velocity of a plane point can be calculated by the relative motion between two keyframes. The LiDAR frames of the two consecutive keyframes are denoted as $L_1$ and $L_2$, and a plane point $\mathbf{P}$ is tracked between the two keyframes, as depicted in Fig. 5 (a). The coordinates of $\mathbf{P}$ in $L_1$ and $L_2$ are $\boldsymbol{p}^{L_1}$ and $\boldsymbol{p}^{L_2}$, respectively, and the velocity of $\mathbf{P}$ can be obtained by

$$\boldsymbol{v}_p = \boldsymbol{v}_p^{L_1} = \boldsymbol{v}_p^{L_2} = \frac{\boldsymbol{p}^{L_2} - \boldsymbol{p}^{L_1}}{\Delta t}, \tag{20}$$

where $\Delta t$ is the time interval. However, the points tracked in the two keyframes may not be the same due to the non-repetitive scanning pattern and point-cloud downsampling. Hence, when conducting LSPC tracking, the distances between same-plane points are constrained within a threshold (0.1 m) to ensure the accuracy of the point velocity calculated by Equation (20), as shown in Fig. 5 (b).

For $S_j = \{\boldsymbol{p}_k^l \in \mathbb{R}^3 \mid k \in \mathbb{N}_j\}$, $\boldsymbol{p}_k^l$ belongs to the keyframe $KF_k$, and its velocity is $\boldsymbol{v}_{p_k}$. The estimated LITD of $KF_k$ is $\hat{t}_{d_k}$, and the estimated $KF_k$ pose $\{\hat{\mathbf{R}}_{b_k}^w, \hat{\boldsymbol{p}}_{b_k}^w\}$ is the IMU pose at $t_{IMU_k} = t_{LiDAR_k} + \hat{t}_{d_k}$. Specifically, the estimated LITD of the current keyframe $KF_N$ is denoted as $\hat{t}_d = \hat{t}_{d_N}$. The difference between $\hat{t}_{d_k}$ and $\hat{t}_d$ is employed to compensate for $\boldsymbol{p}_k^l$ as follows:

$$\hat{\boldsymbol{p}}_k^l = \boldsymbol{p}_k^l - \boldsymbol{v}_{p_k}(\hat{t}_d - \hat{t}_{d_k}). \tag{21}$$

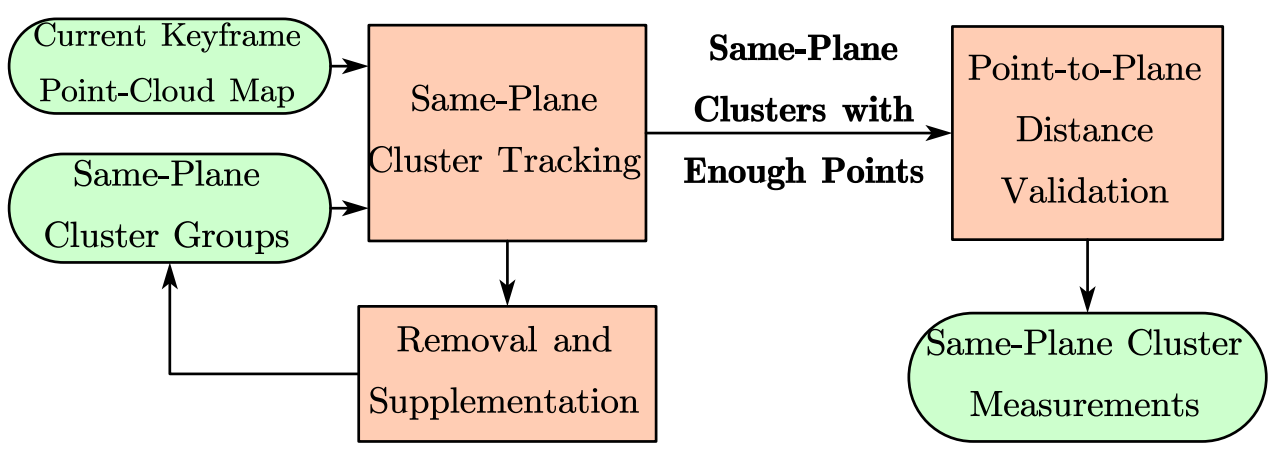

Fig. 4. The overview of the same-plane cluster tracking method.

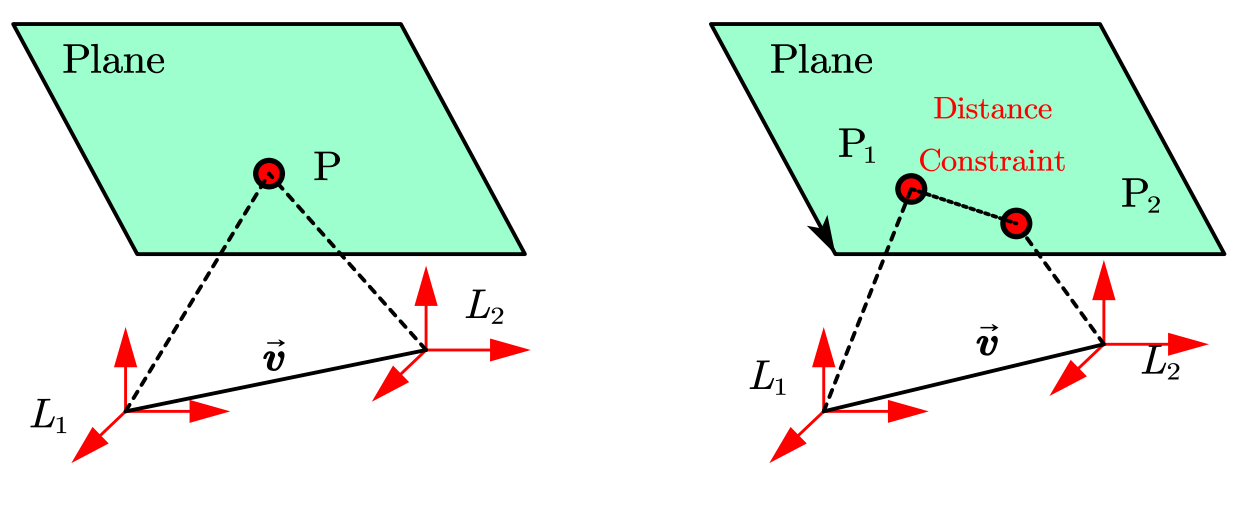

(a)The same point. (b)Downsampled points.

Fig. 5. Point velocity calculation.

Then, $\hat{p}_k^l$ is employed to construct the LSPC measurement by formula (8) and (10). The Jacobian w.r.t the LITD error is

$$\frac{\partial \boldsymbol{r}^j}{\partial \delta t_d} = -\sum_k \hat{\mathbf{J}}_k^j \hat{\mathbf{R}}_{b_k}^w \hat{\mathbf{R}}_l^b \boldsymbol{v}_{p_k} . \tag{22}$$

## V. Experiments and Results

### A. Implementation and Datasets

We implement the proposed MSC-LIO in C++ and Robots Operating System (ROS). We compare the proposed MSC-LIO with other SOTA LIOs, including LIO-SAM (without loop closure) [11], FAST-LIO2 [18] and FF-LINS [12]. LIO-SAM and FF-LINS are optimization-based systems, while FAST-LIO2 is a filter-based system. The LiDAR-IMU extrinsic parameters and time delay are assumed unknown for all systems. The sliding window size for MSC-LIO and FF-LINS are all set to 10 to ensure a fair comparison. All systems are implemented on a desktop PC (AMD R7-3700X).

The used public datasets are the *MCD* [34] and *WHU-Helmet* [35] datasets. Six *KTH* sequences of the *MCD* dataset are selected, with a total trajectory length of 6786 m. The *KTH* sequences were collected using a handheld setup, which was equipped with a spinning LiDAR, i.e., Ouster OS1, and a MEMS IMU. The *WHU-Helmet* dataset was collected by a helmet-based system, which was equipped with a solid-state LiDAR, i.e., Livox AVIA, and a MEMS IMU. In the *WHU-Helmet* dataset, four sequences collected by Livox AVIA are employed, with a total trajectory length of 3656 m. Both the *MCD* and *WHU-Helmet* datasets provide high-precision reference truth.

To further evaluate MSC-LIO in real-world scenarios, we also conducted real-world experiments on a wheeled robot with a maximum speed of 1.5 m/s, as depicted in Fig. 6. The sensors used include a solid-state LiDAR, i.e., Livox Mid-70, and a MEMS IMU ADIS16465. The sensors are well-synchronized

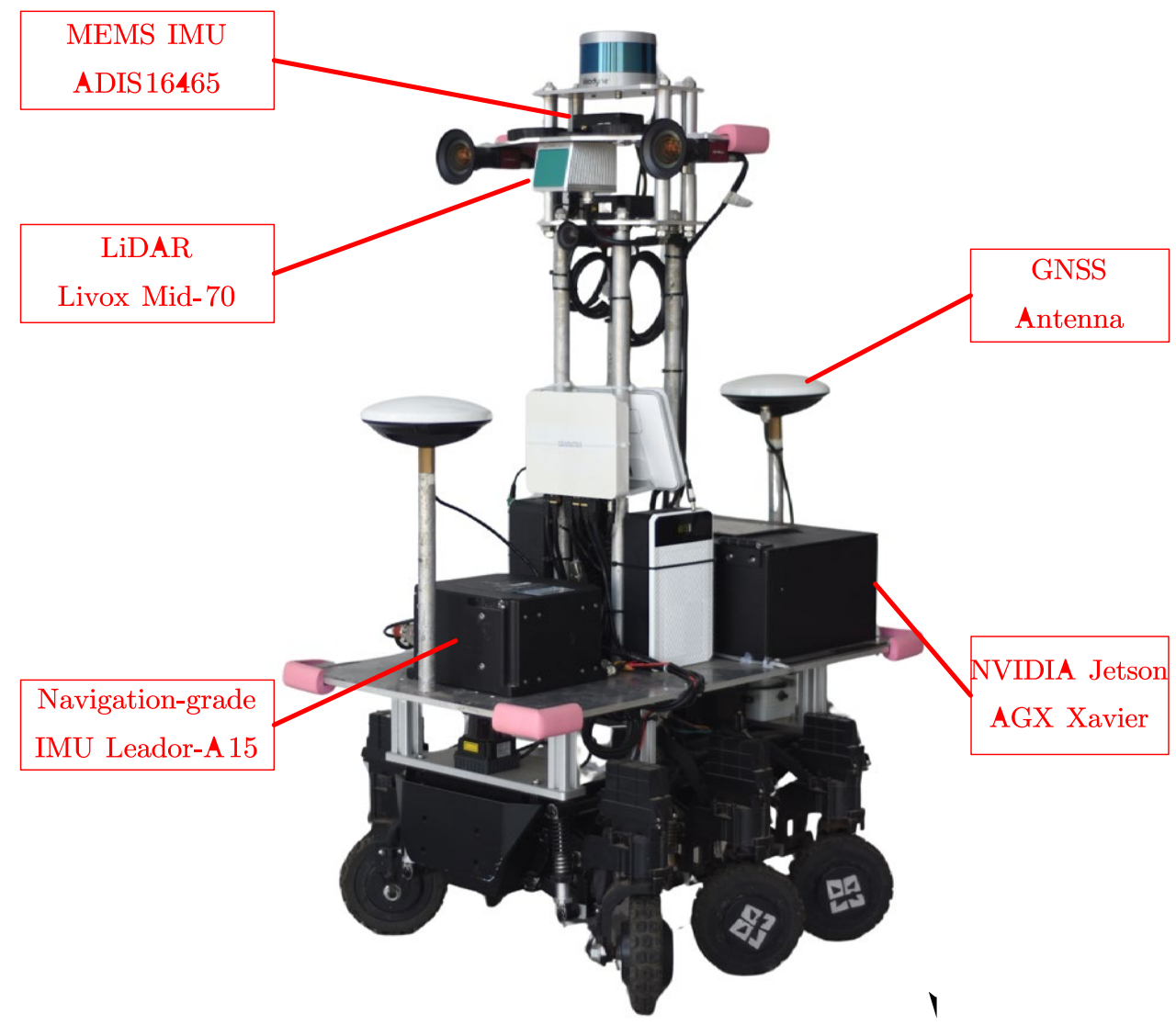

Fig. 6. The robot used for real-world experiments.

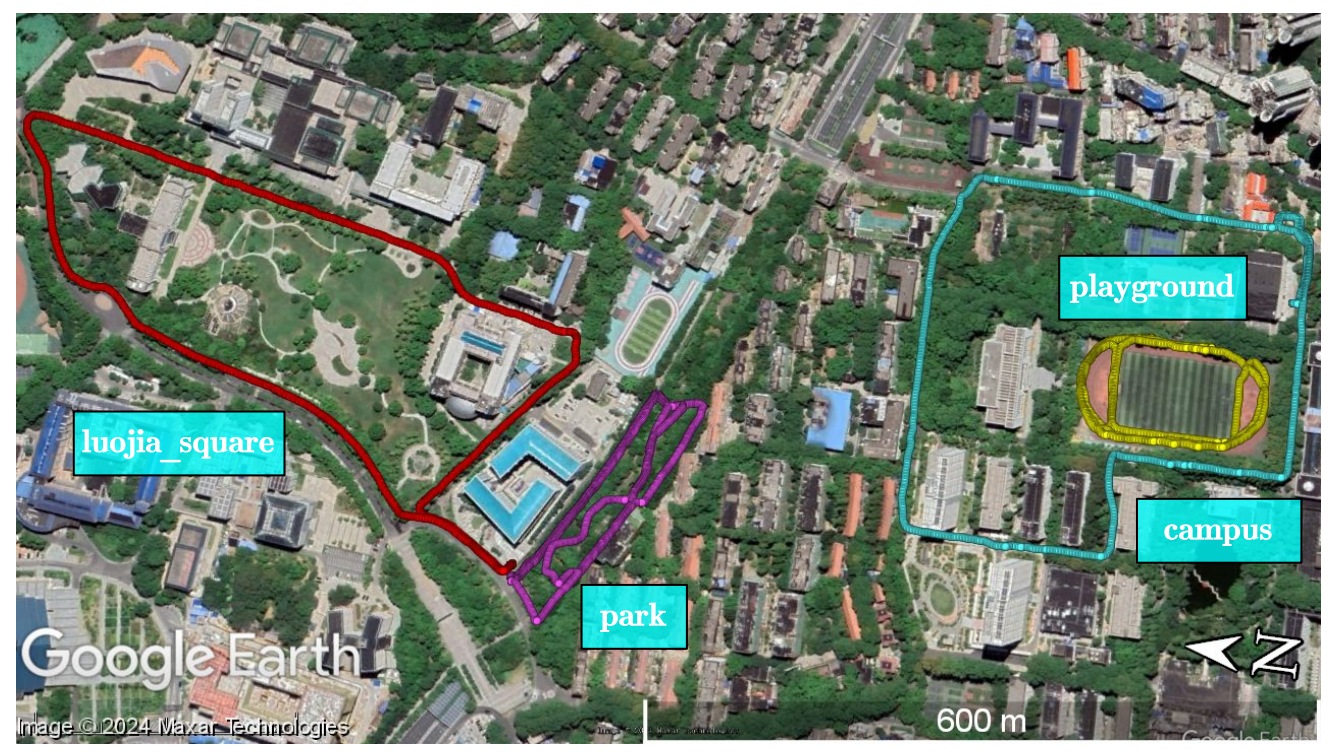

Fig. 7. Four scenes of the real-world experiments. Different colors denote different sequences.

TABLE I
Datasets Descriptions

| Dataset | *MCD* | *WHU-Helmet* | *RobNav* |
|---|---|---|---|
| LiDAR Type | Spinning | Solid-state | Solid-state |
| LiDAR Rate | 10Hz | 10Hz | 10Hz |
| LiDAR Line | 64 | 6 | 1 |
| IMU Rate | 400Hz | 600Hz | 200Hz |
| Total Length | 6786 m | 3656 m | 15248 m |

by using the pulse signal from the GNSS receiver. The integrated navigation solution of a navigation-grade [31] IMU Leador-A15 and high-precision GNSS real-time kinematic (RTK) serves as the ground truth, with position and attitude accuracy of 0.02 m and 0.01 deg, respectively [36]. Eight sequences with a total length of 15248 m were collected by the wheeled robot and denoted as the *RobNav* dataset. The eight experimental scenes include various structured and unstructured environments. In addition, dynamic objects, such as pedestrians, bicycles, and vehicles, make it challenging to achieve robust positioning. Four sequences among them are shown in Fig. 7. The information on the public datasets and the real-world Robnav dataset is summarized in TABLE I. The keyframe selection thresholds primarily depend on the dead-reckoning performance of the IMU and the carrier dynamics to ensure accurate LiDAR F2F data association. For

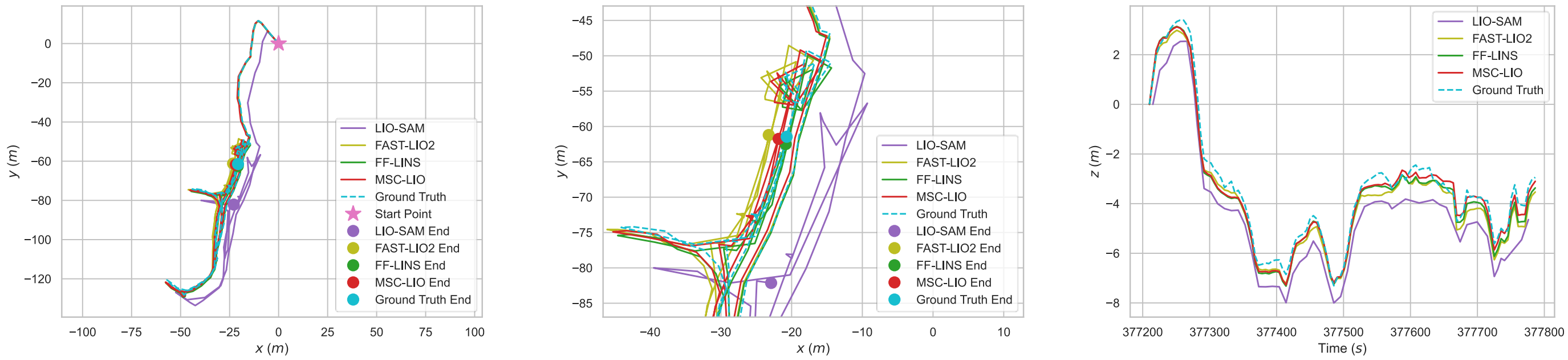

(a) The whole trajectory. (b) The endpoint. (c) The height (z-axis) change.

Fig. 8. Results on the *WHU-Helmet residence* sequence.

TABLE II
ATES ON THE *MCD* DATASET

| ATE (m) | LIO-SAM | FAST-LIO2 | FF-LINS | MSC-LIO |
|---|---|---|---|---|
| *kth_day_06* | 0.84 | 0.46 | 0.57 | **0.27** |
| *kth_day_09* | 1.09 | **0.17** | 0.83 | 0.34 |
| *kth_day_10* | 1.22 | **0.41** | 0.51 | 0.45 |
| *kth_night_01* | 13.19 | 0.52 | 1.12 | **0.33** |
| *kth_night_04* | 0.47 | **0.20** | 0.38 | 0.29 |
| *kth_night_05* | 0.95 | 0.41 | 0.27 | **0.25** |
| RMS | 5.46 | 0.38 | 0.67 | **0.33** |

TABLE III
ATES ON THE *WHU-HELMET* DATASET

| ATE (m) | LIO-SAM | FAST-LIO2 | FF-LINS | MSC-LIO |
|---|---|---|---|---|
| *mall* | 0.55 | **0.32** | 0.69 | 0.48 |
| *residence* | **0.35** | 1.03 | 0.43 | 0.45 |
| *street* | 1.06 | 0.90 | 0.97 | **0.60** |
| *subway* | 28.43 | 2.39 | 2.34 | **1.82** |
| RMS | 14.23 | 1.39 | 1.33 | **1.01** |

instance, the translation thresholds differ between the *WHU-Helmet* dataset and the *RobNav* dataset, with the former set to 0.3 m and the latter to 0.4 m.

## *B. Evaluation of the Accuracy*

1) *Public MCD Dataset*

The absolute translation errors (ATEs) are calculated using evo [37], as presented in TABLE II. LIO-SAM has a significant error on the *kth_nigth_01* sequence, mainly due to its inability to extract effective features in small indoor scenes. Due to the presence of many loop closures in the *MCD* dataset, FAST-LIO2 based on F2M association can establish associations with the self-built map, resulting in better accuracy compared to FF-LINS based on F2F association. MSC-LIO constructs F2F multi-state constraints and thus achieves higher accuracy compared to FF-LINS. Besides, MSC-LIO achieves the best accuracy on three sequences and the smallest root-mean-square (RMS) error on all six sequences.

2) *Public WHU-Helmet Dataset*

The ATEs are calculated on the *WHU-Helmet* dataset, as shown in TABLE III. LIO-SAM nearly fails in the indoor scene of the *subway* sequence, resulting in a great ATE. Benefiting from the multi-state constraints of LSPC measurements, the proposed MSC-LIO achieves the highest accuracy in the *street* and *subway* sequences. Moreover, the ATE RMS of the four sequences indicates that MSC-LIO outperforms other SOTA systems in terms of accuracy.

TABLE IV
ARES AND ATES ON THE *ROBNAV* DATASET

| ARE / ATE (deg / m) | FAST-LIO2 | FF-LINS | MSC-LIO |
|---|---|---|---|
| *campus* | 3.55 / 4.42 | **0.41** / 1.51 | 0.50 / **1.01** |
| *building* | 3.13 / 3.12 | 0.65 / 1.90 | **0.40** / **1.03** |
| *playground* | 2.84 / 1.59 | 0.77 / 1.27 | **0.55** / **0.75** |
| *park* | 3.24 / 4.00 | **0.90** / 1.44 | 0.97 / **1.25** |
| *cs_campus* | 3.68 / 4.38 | 0.93 / 2.04 | **0.84** / **1.67** |
| *luojia_square* | 3.47 / 5.18 | 0.88 / 3.88 | **0.22** / **0.92** |
| *east_lake* | 3.20 / 4.49 | 1.48 / 8.39 | **0.88** / **3.28** |
| *library* | 3.28 / 2.73 | **0.37** / **1.77** | 0.39 / 1.93 |
| RMS | 3.31 / 3.90 | 0.86 / 3.58 | **0.65** / **1.67** |

The results of the *residence* sequence are shown in Fig. 8. In Fig. 8 (a), the whole trajectory of LIO-SAM exhibits a significant drift compared to the ground truth trajectory, while the trajectories of other systems are well aligned with the truth. Fig. 8 (b) and (c) show that near the endpoint, FF-LINS and MSC-LIO are closer to the ground truth than FAST-LIO2, which benefits from the consistent F2F association. MSC-LIO exhibits a smaller error than FF-LINS, especially in the z-axis, as MSC-LIO constructs accurate multi-state constraints.

3) *Real-World Experiments*

The absolute rotation errors (AREs) and ATEs are calculated, as presented in TABLE IV. LIO-SAM fails to run on the real-world experiments, i.e., the *RobNav* dataset, because the used Livox Mid-70 point clouds are sparse and difficult to extract features effectively. In contrast, other systems do not explicitly extract features, and thus they can run successfully. The AREs of FAST-LIO2 are mostly larger than 3 degrees because it cannot estimate the LiDAR-IMU extrinsic parameters. In contrast, FF-LINS and MSC-LIO can estimate the extrinsic parameters, resulting in much smaller AREs. MSC-LIO achieves the highest accuracy in most sequences, and the RMSs of ATEs and AREs are both at optimal levels.

The results of the *luojia_square* sequence are shown in Fig. 9. Fig. 9 (a) and (b) show that the trajectory of MSC-LIO is better aligned with the ground truth, while FF-LINS and FAST-LIO2 exhibit significant drifts in the horizontal and elevation directions, respectively. The attitude error is shown in Fig. 9 (c). FAST-LIO2 has large errors in the roll, pitch, and yaw angles due to the F2M association that may introduce incorrect observability. With consistent F2F association, FF-LINS and MSC-LIO exhibit high accuracy in the observable roll and pitch

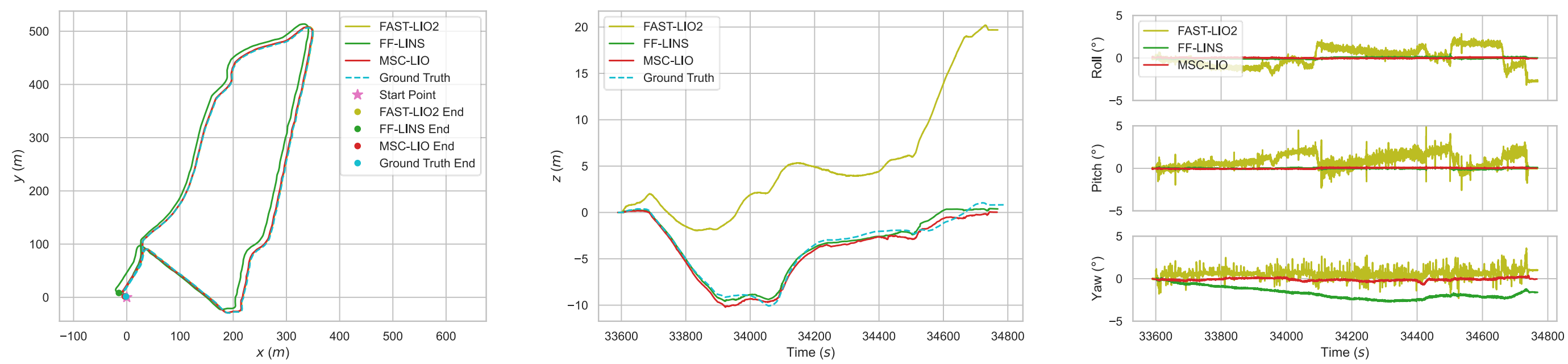

(a) The whole trajectory. (b) The height (z-axis) change. (c) The attitude error.

Fig. 9. Results on the *RobNav luojia_square* sequence.

TABLE V
THE AVERAGE RUNNING TIME PER KEYFRAME AND THE TOTAL TIME OF DATA ASSOCIATION AND STATE ESTIMATION

| | FAST-LIO2 | | | FF-LINS | | | MSC-LIO w/o Tracking | | | MSC-LIO | | |
|---|---|---|---|---|---|---|---|---|---|---|---|---|
| | $t_{\mathrm{DA}}$ (ms) | $t_{\mathrm{Est}}$ (ms) | $t_{\mathrm{Total}}$ (s) | $t_{\mathrm{DA}}$ (ms) | $t_{\mathrm{Est}}$ (ms) | $t_{\mathrm{Total}}$ (s) | $t_{\mathrm{DA}}$ (ms) | $t_{\mathrm{Est}}$ (ms) | $t_{\mathrm{Total}}$ (s) | $t_{\mathrm{DA}}$ (ms) | $t_{\mathrm{Est}}$ (ms) | $t_{\mathrm{Total}}$ (s) |
| *campus* | 5.2 | 0.7 | 65 | 17.1 | 38.7 | 198 | 15.3 | 10.8 | 92 | 5.5 | 11.2 | 59 |
| *building* | 5.1 | 0.7 | 106 | 15.8 | 42.1 | 58 | 14.8 | 11.0 | 153 | 5.0 | 10.6 | 92 |
| *playground* | 7.8 | 0.9 | 84 | 17.5 | 46.6 | 194 | 14.7 | 8.7 | 71 | 8.0 | 12.9 | 63 |
| *park* | 4.6 | 0.6 | 65 | 17.1 | 34.1 | 169 | 14.5 | 11.3 | 84 | 4.0 | 9.6 | 45 |
| *cs_campus* | 4.9 | 0.6 | 102 | 15.9 | 41.0 | 140 | 13.5 | 9.3 | 134 | 5.1 | 10.0 | 89 |
| *luojia_square* | 5.0 | 0.7 | 68 | 17.8 | 40.7 | 209 | 14.5 | 10.9 | 91 | 5.5 | 11.1 | 59 |
| *east_lake* | 4.3 | 0.7 | 72 | 15.9 | 32.4 | 269 | 12.7 | 8.8 | 109 | 3.5 | 7.9 | 59 |
| *library* | 5.2 | 0.7 | 95 | 18.3 | 42.7 | 316 | 15.2 | 11.0 | 135 | 5.5 | 11.3 | 87 |
| Average | 5.3 | 0.7 | 82 | 16.9 | 39.8 | 194 | 14.4 | 10.2 | 108 | 5.3 | 10.6 | 69 |

Here, $t_{\mathrm{DA}}$ and $t_{\mathrm{Est}}$ represents the average running time of data association and state estimation for each keyframe, respectively; $t_{\mathrm{Total}}$ represents the total time of data association and state estimation for the entire sequence.

angles, though the yaw angle may diverge for unobservability. MSC-LIO exhibits higher accuracy in the yaw angle with a maximum error of 0.72 degrees because of the multi-state constraints of LSPC measurements.

## C. *Evaluation of the Efficiency*

We compared the efficiency of MSC-LIO with FAST-LIO2, FF-LINS, and MSC-LIO w/o Tracking on the *RobNav* dataset. MSC-LIO w/o Tracking does not use the proposed LSPC tracking; instead, its data association works as follows. The current keyframe point clouds are sampled and voxel downsampled. Then, the obtained points are projected onto each historical keyframe point-cloud map within the sliding window. Finally, planes are fitted by searching for neighboring points, and the nearest neighboring points from all keyframes are employed to construct the LSPC measurements. For more details, please refer to BA-LINS [17]. The running time of data association and state estimation is recorded, as shown in TABLE V.

### 1) *Efficiency of Data Association*

Data association includes F2F or F2M feature association. For MSC-LIO, data association consists of the preprocessing of LSPCs (removal and supplementation) and LSPC tracking. FAST-LIO2 and MSC-LIO spend less time in data association according to TABLE V. The reason is FF-LINS and MSC-LIO w/o Tracking perform nearest neighboring point search and plane fitting with all historical keyframes within the sliding window. In contrast, FAST-LIO2 only matches the current frame with the map, and MSC-LIO only matches the latest points in the LSPCs with the current keyframe. Although the preprocessing of LSPCs consumes extra time, the data association efficiency of MSC-LIO is still significantly improved, with the time spent approximately one-third of MSC-LIO w/o Tracking.

In fact, the data association time of MSC-LIO is shorter than FAST-LIO2 for the whole sequences, as MSC-LIO only processes the LiDAR frames at each keyframe moment, while FAST-LIO2 performs F2M association at each LiDAR frame moment. Due to the keyframe selection strategy, the average keyframe interval is 3-4 times of LiDAR frames on the *RobNav* dataset. This is why the running time for single data association and state estimation of MSC-LIO is longer than FAST-LIO2, but the total running time of MSC-LIO is shorter in TABLE V.

### 2) *Efficiency of State Estimation*

State estimation includes the prediction and update, or factor graph optimization. For MSCKF, it also includes extra state augmentation and marginalization. As shown in TABLE V, the state estimation time of the optimization-based FF-LINS is much longer than other filter-based methods. The state estimation time of MSC-LIO w/o Tracking and MSC-LIO is similar, as the number of plane features is close. The state estimation efficiency of FAST-LIO2 is higher than MSCKF-based methods, as its state vector dimension is lower. In contrast, the state vectors of MSCKF-based methods contain the keyframe states within the sliding window, and extra state augmentation and marginalization are also required.

## D. *Ablation Experiments*

### 1) *The Impact of the Same-Plane Cluster Tracking*

In Section V.C, it has been demonstrated that the LSPC

TABLE VI
IMPACT OF SAME-PLANE CLUSTER TRACKING ON THE NUMBER OF PLANES, AREs AND ATEs

| | MSC-LIO w/o Tracking | | MSC-LIO | |
|---|---|---|---|---|
| | plane number | ARE / ATE (deg / m) | plane number | ARE/ATE (deg / m) |
| *campus* | 350 | **0.36** / 1.15 | **356** | 0.50 / **1.01** |
| *building* | **361** | 0.85 / 1.97 | 341 | **0.40** / **1.03** |
| *playground* | 267 | 0.75 / 0.88 | **390** | **0.55** / **0.75** |
| *park* | **369** | 1.34 / 2.18 | 291 | **0.97** / **1.25** |
| *cs_campus* | 305 | **0.52** / **1.33** | **313** | 0.84 / 1.67 |
| *luojia_square* | **360** | 0.38 / 1.97 | 343 | **0.22** / **0.92** |
| *east_lake* | **285** | 1.03 / 5.34 | 242 | **0.88** / **3.28** |
| *library* | **360** | 0.53 / **1.81** | 359 | **0.39** / 1.93 |
| RMS | **334** | 0.79 / 2.45 | 332 | **0.65** / **1.67** |

TABLE VII
IMPACT OF THE TIME-DELAY ESTIMATION METHODS ON ATEs

| ATE (m) | MSC-LIO w/o TD | MSC-LIO-PTD | MSC-LIO |
|---|---|---|---|
| *kth_day_06* | 0.78 | **0.25** | 0.27 |
| *kth_day_09* | 1.14 | 0.36 | **0.34** |
| *kth_day_10* | 2.65 | 0.46 | **0.45** |
| *kth_night_01* | 1.60 | **0.33** | **0.33** |
| *kth_night_04* | 1.13 | 0.35 | **0.29** |
| *kth_night_05* | 0.71 | 0.26 | **0.25** |
| RMS | 1.49 | 0.34 | **0.33** |

tracking improves the efficiency of data association. In this part, we will further evaluate its impact on accuracy. The statistical results of the number of planes, AREs, and ATEs on the *RobNav* dataset are shown in TABLE VI. The number of planes refers to the average number of LSPCs selected as valid measurements at each keyframe selection moment. The proposed LSPC tracking method improves the accuracy, with an average reduction of 17.7% in AREs and 31.8% in ATEs, while the average number of planes remains relatively unchanged. The improvement in accuracy stems not from the change in the number of planes but from the more uniform distribution of the tracked same-plane points.

2) *The Impact of the Proposed LiDAR-IMU Time Delay Estimation Method*

In this section, the impact of the LITD estimation method on accuracy is evaluated. The traditional LITD estimation methods use the LITD to compensate for keyframe poses, as detailed in [38]. MSC-LIO with the traditional LITD estimation method is denoted as MSC-LIO-PTD, and MSC-LIO without LITD estimation is denoted as MSC-LIO w/o TD. The *MCD* dataset has a relatively larger LITD, while the LITDs in the *WHU-Helmet* and *RobNav* datasets are less than 2 ms. Hence, the *MCD* dataset is employed to evaluate the effect of LITD estimation methods.

Fig. 10 depicts the estimated LITDs on the *kth_day_06* and *kth_night_01* sequence. The estimated LITDs are close and have converged, verifying the feasibility of the proposed LITD estimation method. The ATEs with different LITD estimation methods are shown in TABLE VII. The ATEs increase greatly when the large LITD is not estimated. This is mainly because the LITD causes inaccuracy in the IMU poses used for point cloud distortion correction, resulting in low accuracy of LSPC measurements. The ATEs of MSC-LIO-PTD and MSC-LIO are similar for all sequences, indicating that the accuracy of the proposed point-velocity-based LITD estimation method is comparable to the traditional method. However, the proposed method is much simpler and more direct, as it employs only point velocity to compensate for the LiDAR points rather than using both the angular rate and velocity to compensate for all keyframe poses within the sliding window.

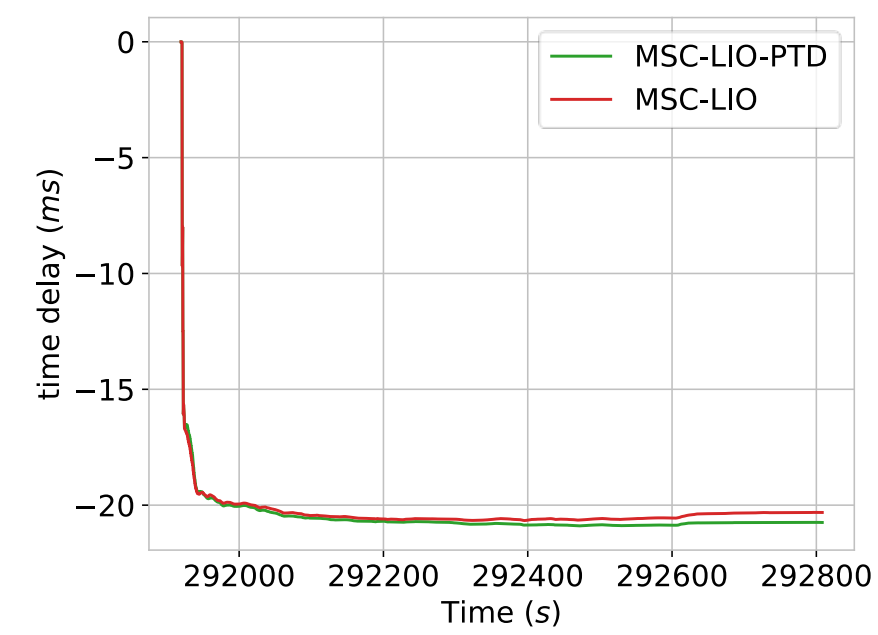

(a) *kth_day_06*

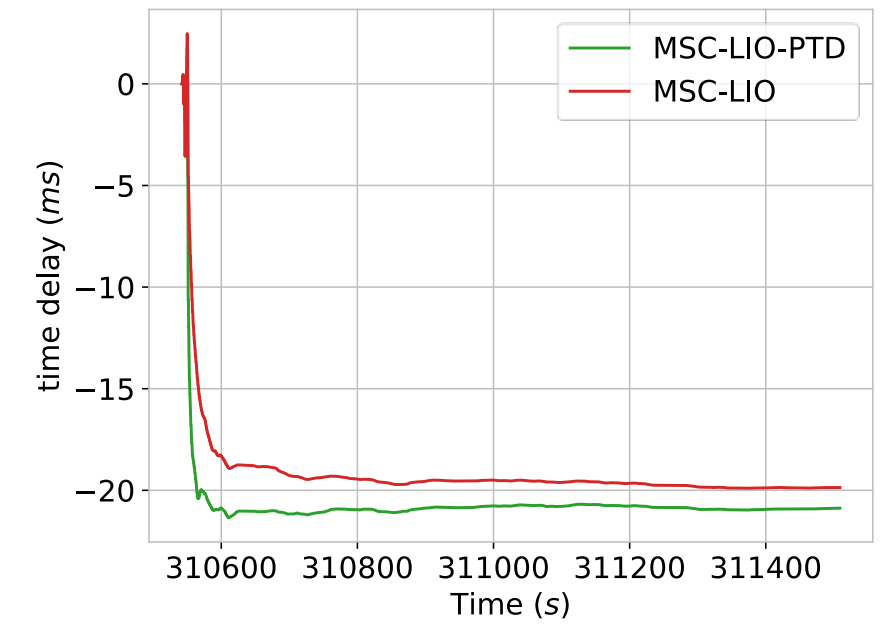

(b) *kth_night_01*

Fig. 10. The LiDAR-IMU time delays of MSC-LIO-PTD and MSC-LIO. MSC-LIO-PTD uses the time delay to compensate for the pose of each keyframe.

## E. *Real-Time Performance*

To further test the real-time performance of MSC-LIO, we also implemented the MSC-LIO on an edge device NVIDIA jetson AGX Xavier, as shown in Fig. 6. The AGX Xavier features an 8-core ARM v8.2, 64-bit CPU with 8MB L2 and 4MB L3 cache, along with 16GB of 256-bit LPDDR4x memory running at 2133MHz, providing a bandwidth of 137GB/s. We collected two extra sequences in real-world environments by the wheeled robot shown in Fig. 6. One is the *information_campus* sequence (2012 s and 2565 m), and the other is the *academic_building* sequence (1152 s and 1457 m), as depicted in Fig. 11. The ground truth is obtained from the integrated navigation solution of Leador-A15 and GNSS-RTK as mentioned above.

We run the proposed MSC-LIO on the edge device NVIDIA Jetson AGX Xavier in real-time mode. The estimated trajectories are well-aligned with the ground truth, as shown in Fig. 12. The real-time performance including accuracy and efficiency is shown in TABLE VIII. The small AREs and ATEs indicate that the proposed MSC-LIO exhibits high accuracy when running in real-time mode on the AGX Xavier. The total running times of MSC-LIO are less than one-fifth of the total duration of the sequences. Besides, the average processing time per LiDAR frame is calculated by dividing the total running time by the total number of LiDAR frames (the LiDAR frame rate is 10 Hz in the real-time experiment). On the edge device AGX Xavier, the average processing times per LiDAR frame of the *information_campus* and *academic_building* are 0.02 s and 0.017 s, respectively, exhibiting superior efficiency. In summary, the proposed MSC-LIO demonstrates excellent real-time performance on the edge device NVIDIA Jetson AGX Xavier.

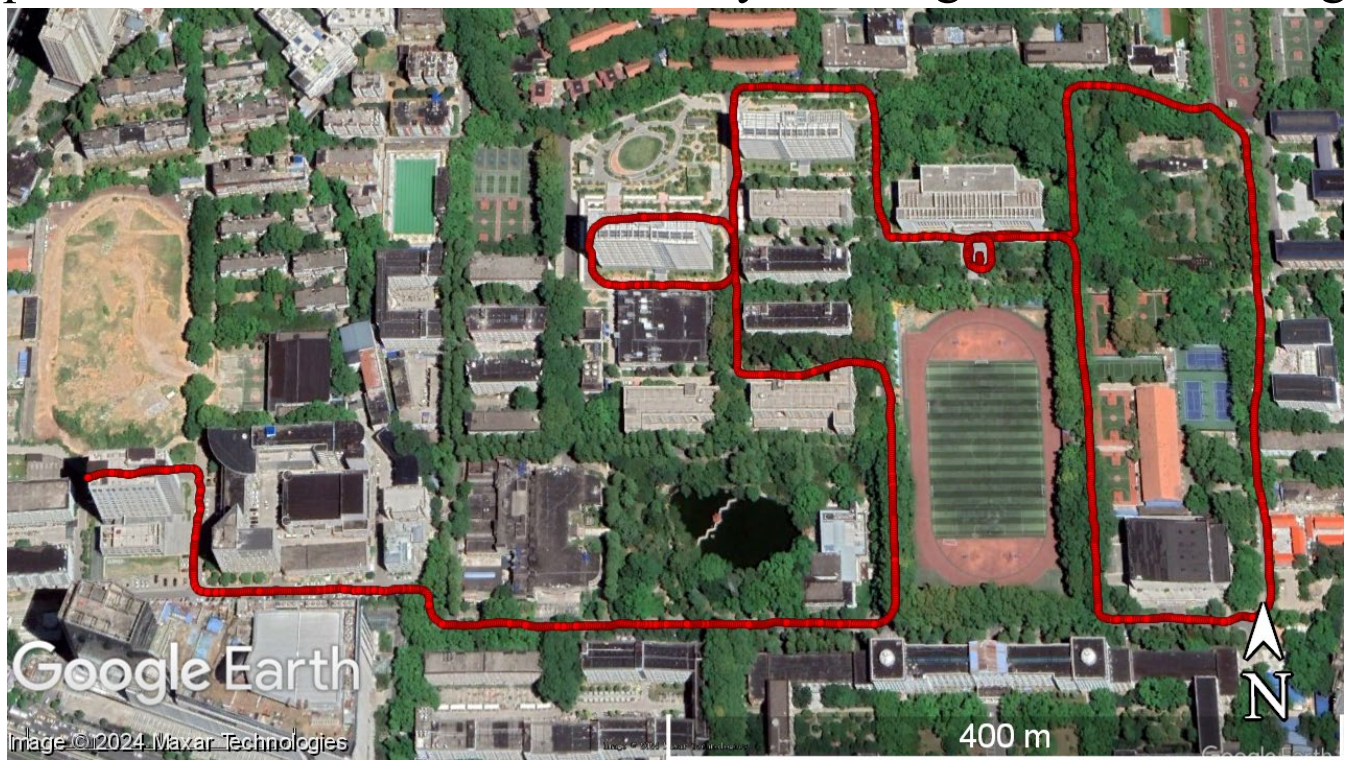

(a) *information_campus*

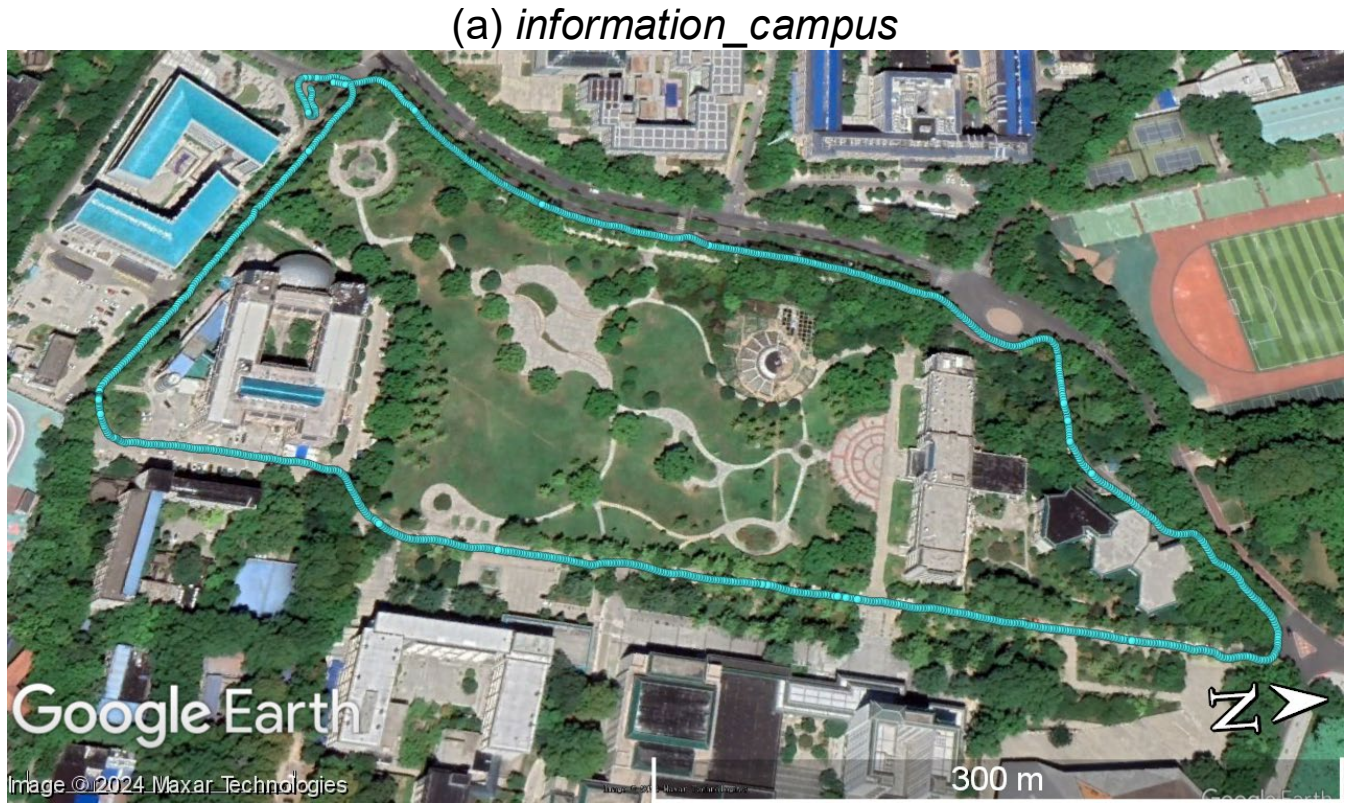

(b) *academic_building*

Fig. 11. The testing scenes of the real-time experiment.

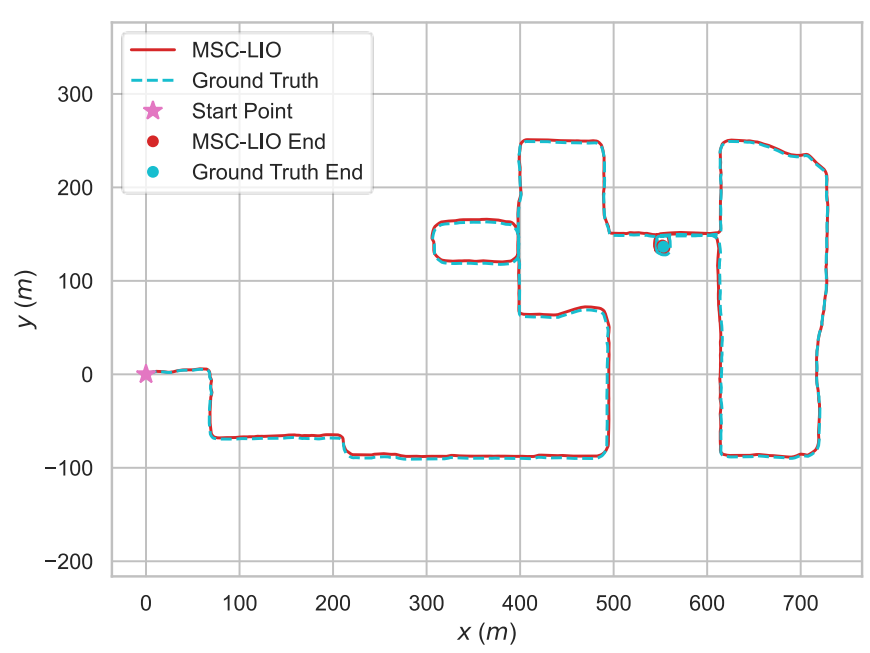

(a) *information_campus*

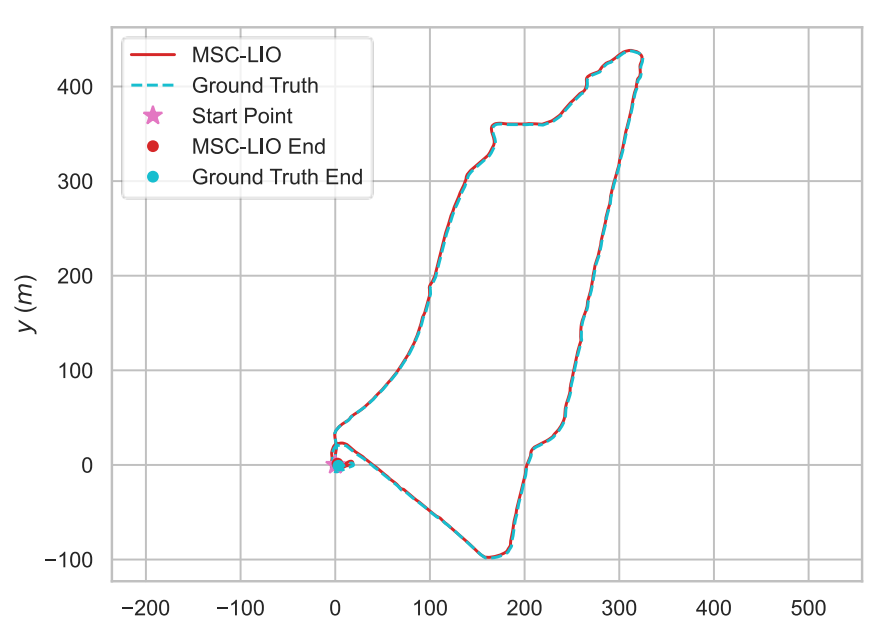

(b) *academic_building*

Fig. 12. Estimated trajectories on the real-time experiments.

TABLE VIII
REAL-TIME PERFORMANCE OF MSC-LIO

| sequence | ARE / ATE (deg / m) | total running time (s) | average processing time per frame (s) |
|---|---|---|---|
| *information_campus* | 0.50 / 1.47 | 393 | 0.020 |
| *academic_building* | 0.49 / 0.73 | 196 | 0.017 |

## VI. CONCLUSION

This paper proposes a tightly coupled LIO within the MSCKF framework. An F2F same-plane cluster tracking method is designed to improve data association efficiency, and the same-plane cluster measurement model constructs a multi-state constraint. In addition, we propose a simpler and more direct point-velocity-based LiDAR-IMU time-delay estimation method based on the same-plane cluster tracking. We conducted extensive experiments on both public datasets and real-world environments. The experimental results demonstrate that the proposed MSC-LIO outperforms SOTA systems in terms of accuracy and efficiency. Future work involves integrating the absolute positioning sources, such as the GNSS and the UWB, to achieve drift-free localization in large-scale outdoor and indoor environments.

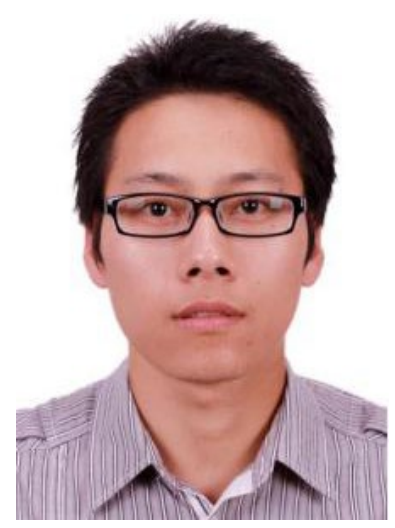

**Tisheng Zhang** received the B.Sc. and Ph.D. degrees in communication and information systems from Wuhan University, Wuhan, China, in 2008 and 2013, respectively.

From 2018 to 2019, he was a Postdoctoral Researcher with The Hong Kong Polytechnic University, Hong Kong. He is an Associate Professor with the GNSS Researches Center, Wuhan University. His research interests include GNSS receiver and multi-sensor deep integration.

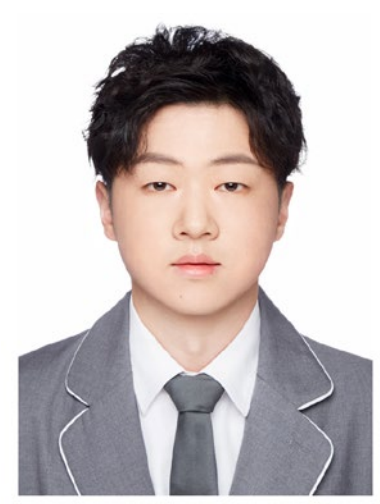

**Man Yuan** received the B.E. (Hons.) degree in communication engineering from Wuhan University, Wuhan, China, in 2023, where he is currently pursuing the master's degree in navigation, guidance, and control with the GNSS Research Center.

His primary research interests include GNSS/INS integrations and LiDAR-based navigation.

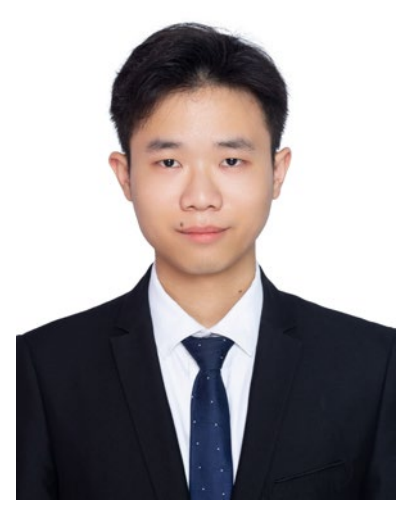

**Linfu Wei** received the B.E. and M.E. (Hons.) degrees from Wuhan University, Wuhan, China, in 2021 and 2024, respectively.

His primary research interests include LiDAR-inertial odometry, UWB-based navigation, and multi-sensor fusion.

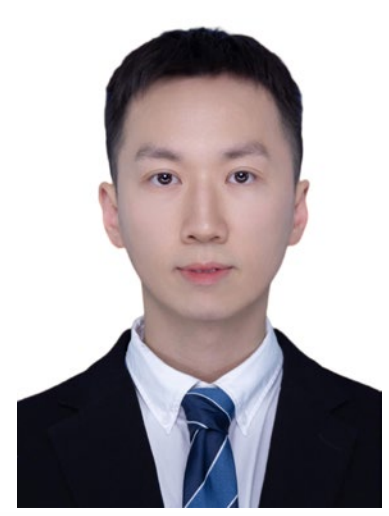

**Hailiang Tang** received the M.E. and Ph.D. degrees from Wuhan University, Wuhan, China, in 2020 and 2023, respectively.

He is currently a Postdoctoral Fellow with the GNSS Research Center, Wuhan University. His current research interests include visual and LiDAR SLAM, autonomous robotics systems, GNSS/INS integration technology, and deep learning.

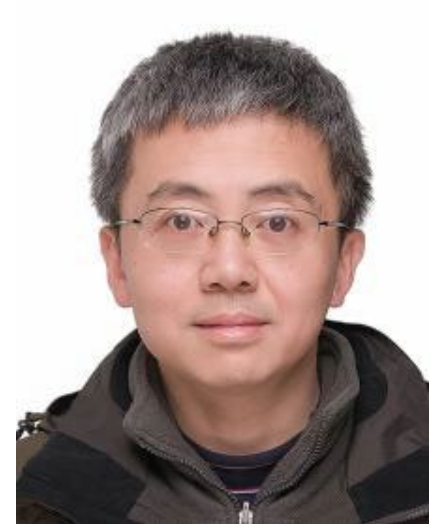

**Xiaoji Niu** received the bachelor's and Ph.D. degrees from the Department of Precision Instruments, Tsinghua University, Beijing, China, in 1997 and 2002, respectively.

He was a Postdoctoral Researcher with the University of Calgary, Calgary, AB, Canada, and worked as a Senior Scientist with SiRF Technology Inc., Shanghai, China. He is currently a Professor with the GNSS Research Center, Wuhan University, Wuhan, China. He has authored or coauthored more than 90 academic articles and owns 28 patents. He leads a multi-sensor navigation group focusing on GNSS/INS integration, low-cost navigation sensor fusion, and its new applications.